\RequirePackage{fix-cm}
\documentclass[smallcondensed]{svjour3}      
\smartqed  
\usepackage{xcolor}
\usepackage{graphicx}
\usepackage{url}
\usepackage{amsmath}
\usepackage{amssymb}
\usepackage{booktabs}
\usepackage{caption}
\usepackage[linesnumbered,ruled]{algorithm2e}
\journalname{Journal (2019)}
\begin{document}

\title{Patch augmentation: Towards efficient decision boundaries for neural networks
}

\titlerunning{Patch Augmentation}

\author{Marcus D. Bloice \and Peter M. Roth \and Andreas Holzinger}


\institute{Marcus D. Bloice \and Andreas Holzinger
          \at Institute for Medical Informatics, Statistics, and Documentation, Medical University Graz, Austria \\
          \email{marcus.bloice@medunigraz.at}
          \and
          Peter M. Roth
          \at Institute of Computer Graphics and Vision, Graz University of Technology, Austria
}

\date{}

\maketitle

\begin{abstract}
In this paper we propose a new augmentation technique, called \textit{patch augmentation}, that, in our experiments, improves model accuracy and makes networks more robust to adversarial attacks. In brief, this data-independent approach creates new image data based on image/label pairs, where a patch from one of the two images in the pair is superimposed on to the other image, creating a new augmented sample. The new image's label is a linear combination of the image pair's corresponding labels. Initial experiments show a several percentage point increase in accuracy on CIFAR-10, from a baseline of approximately 81\% to 89\%. CIFAR-100 sees larger improvements still, from a baseline of 52\% to 68\% accuracy. Networks trained using patch augmentation are also more robust to adversarial attacks, which we demonstrate using the Fast Gradient Sign Method.

\keywords{Augmentation \and Adversarial attacks \and Decision boundaries}
\end{abstract}

\section{Introduction}
An adversarial misclassification, or adversarial attack, occurs when an image that should seemingly be easily classified correctly by a neural network is suddenly classified as belonging to a completely different class---and with high confidence. Such occurrences are difficult to diagnose and are a cause of much concern in artificial intelligence research, as any model trained with empirical risk minimisation seems to be vulnerable to such attacks. The ease at which neural networks are susceptible to adversarial perturbations are partially the result of images lying close to the decision boundaries that are typically learned by neural networks during their training. Patch augmentation is an attempt to train more efficient decision boundaries.

A recent article in \textit{Nature} has reported on the concerns within artificial intelligence research at the ease at which neural networks are ``fooled'' by adversarial examples \cite{heaven2019deep}. In other words, it has increasingly been shown that state-of-the-art neural networks are failing in quite unexpected and often catastrophic ways. An adversarial example is data, for example an image, that contains an almost imperceptible alteration to it, which when fed in to a neural network causes it to misclassify the image in unpredictable ways, often with a high degree of confidence. This is believed to be at least partially caused by the decision boundaries that are typically learned by neural networks in combination with empirical risk minimisation.

The term ``adversarial examples'' was, for instance, discussed in \cite{szegedy2013adversarial} where they were defined as images that were able to ``trick'' well performing neural networks, seemingly with ease. This led to concerns that neural networks were perhaps brittle or fragile, and that despite counter measures such as regularisation, networks were in fact memorising data sets too closely. The fragility of neural networks to adversarial attacks is perhaps most clearly illustrated by work performed in \cite{su2019pixel} that showed networks can output a confident, incorrect prediction by altering just a single pixel in an input image.

The real life seriousness of adversarial examples was demonstrated nicely by \cite{eykholt2018robust}, where the authors demonstrated that road signs that are trivial to correctly identified by the human eye are interpreted incorrectly by a neural network. For instance, a speed limit or stop sign was recognised as a higher speed limit, causing serious, life-threatening dangers.

Neural networks and deep learning are increasingly being used in medicine, particularly in the field of digital pathology where AI is seen as an important and potentially impactful area where highly skilled pathologists could be supported by deep learning-based systems. However, a study has shown that cancer misdiagnosis can occur due to tiny adversarial perturbations, in networks that otherwise perform extremely well \cite{finlayson2019}.

A different, but yet more serious problem is that adversarial examples can be transferred across different networks \cite{szegedy2013adversarial}. Even when trained with different hyper-parameters, such as the number of layers or trained on different subsets, the networks are still susceptible to the same adversarial attacks. In other words, one does not need access to the trained network in order to devise an adversarial attack.

Adversarial examples are not limited to neural networks, however. It has been shown that several types of algorithm are susceptible to adversarial misclassifications \cite{papernot2016transferability}). It has been shown that multi-class Logistic Regression, Support Vector Machines, Decision Trees, and $k$-Nearest Neighbours are all susceptible to exactly the same types of attacks or examples.

These issues are have been attributed to images that lie close to the learned decision boundaries in trained neural networks \cite{ghorbani2019interpretation}. Attempts are afoot to regularise training so that more efficient decision boundaries can be learned. In contrast, we tackle the problem by introducing a data augmentation method which coerces the network to taking less confident decisions, making them more robust to adversarial perturbations and  thus  less  prone  to  misclassifications. Therefore, in this paper we propose an augmentation technique to counter the issues that result in unfavourable decision boundaries. Primarily, this to alleviate the issue of adversarial attacks and misclassifications, however the technique is also useful to mitigate against memorisation and can make networks more general.

To demonstrate the benefits of our approach, we show experimental results for two different scenarios. First, we show that using the proposed augmentation technique results in a significant increase in classification accuracy for CIFAR-10 and CIFAR-100 using ResNet v1 and v2. Second, we demonstrate that, at the same time, the trained networks are much more robust against adversarial attacks.

\section{Related Work}
Image augmentation is a technique used to regularise networks and make them more general. The procedure involves expanding an existing data set by applying transformations to the training image data in order to create new data, while ensuring that the data's labels are preserved. Increasing the size of a data set in this way can prevent memorisation of the data set, and hence prevent over-fitting. Image augmentation in computer vision is a widely used technique, and became popular for neural network training since the work of \cite{lecun1998gradient}, where random distortions were applied to the original MNIST images to create new samples. The distortions that can be applied to images range from simple rotations or flips along the horizontal and vertical axes, to more elaborate methods, such as elastic distortions. Indeed, the wide breadth of different types of augmentation has led to the development of stand-alone augmentation libraries, such as \textit{Augmentor}\footnote{See \url{https://github.com/mdbloice/Augmentor}}, written by the first author of this paper \cite{bloice2019augmentorbio}.

However, image augmentation is generally a data-dependent task that requires a certain amount of domain knowledge in order to develop a successful augmentation strategy. For example, it must be clear that any transformations that are applied to the image data are label preserving---for instance, that a horizontal flip will not result in an image where its label is no longer representative of the image's contents. A simple example might be the images in a digit recognition task---the figure 8 can be translated through both the horizontal and vertical axes while preserving the image's label, while the figure 7 cannot be translated through either axis without its label no longer representing the image. Hence, augmentation strategies often require domain knowledge.

Approaches to create data-independent augmentation techniques include AutoAugment \cite{cubuk2018auto} where the augmentation procedure is learned as the network is trained, adjusting the augmentation policy's hyper-parameters on the fly. However, this procedure adds a further learning procedure, which requires more training time and iterations. More recently than AutoAugment, a population-based automatic augmentation technique has been proposed \cite{ho2019population} which performs more efficiently than AutoAugment. Work by \cite{zhang2018} specifically demonstrated augmentation as an approach to the decision boundary problem and adversarial examples. They demonstrated the \textit{mixup} algorithm which performs a linear combination of two images and their labels to create augmented images. This approach proved to be highly effective, and in this work we wish to propose a similar, more parametric technique which outperforms mixup in our initial experiments.

Therefore, in this work, we discuss patch augmentation, which is a data-independent approach aimed at improving model generalisation and mitigating against adversarial attacks. In Section~\ref{sec:implementation} we describe our approach in detail and give experimental results in Section~\ref{sec:experiments}.

\section{How Patch Augmentation Works}\label{sec:howitworks}
\begin{figure}[h]
  \centering
  \includegraphics[width=0.40\textwidth]{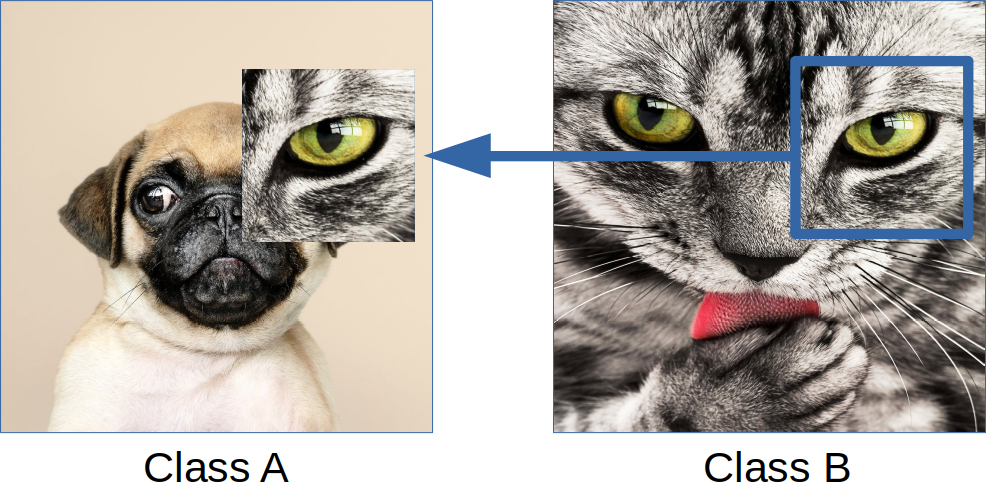}
  \caption{Patch augmentation creates new image data by extracting a patch from an image, in this case from \textit{Class B}, and placing it in to another image from a different class, in this case \textit{Class A}. The location from where the patch is extracted is randomly chosen for each newly created, augmented image. Its new label, which must be one-hot-encoded, is computed from the original image pair's two labels, in this case the computed label is $\tilde{y} = [0.85, 0.15]$, as the patch's area is 15\% of the new image's size and the original image pair's labels are $[1.0, 0.0]$ and $[0.0, 1.0]$ for Class A and Class B respectively.}
  \label{fig:patch-augmentation-main-example}
\end{figure}

In essence, the idea behind patch augmentation is to create new image data based on pairs of images and their labels. As Figure~\ref{fig:patch-augmentation-main-example} shows, a patch of a certain size is copied from image A and superimposed on to image B at a random location, creating a new augmented image. The new image's label, which must be one-hot-encoded, is a linear combination of the image pair's labels, based on the size of the area of the patch in relation to the size of the entire image, in pixels.

\begin{figure}[h]
  \centering
  \includegraphics[width=0.40\textwidth]{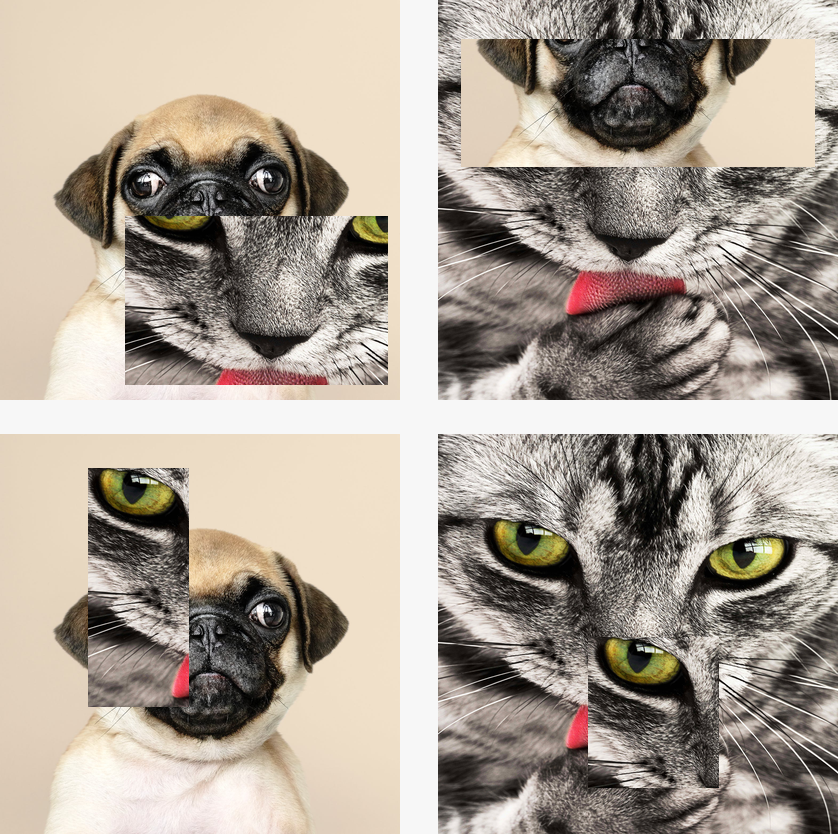}
  \caption{Examples of images created using patch augmentation. Patches are extracted from a random image in the training set, and can be from within the same class, as shown in the bottom right image. Clockwise from the top left, the image labels are [0.72220625, 0.27779375], [0.2832, 0.7168], [0.0, 1.0] and [0.918925, 0.081075] respectively.}
  \label{fig:patch-augmentation-random-patches-example}
\end{figure}

The patches that are extracted are of a random shape and size, the area of which are controlled by minimum/maximum width/height parameters. Figure~\ref{fig:patch-augmentation-random-patches-example} shows a number of examples of images created by patch augmentation for a binary class data set. Notice that the augmented images can contain patches from the same class. In order to demonstrate the method in a fully reproducible manner, source code, Jupyter demonstration notebooks, and the complete experimental setup described in this paper can be found here under:

\begin{center}
\url{https://github.com/mdbloice/Patch-Augmentation}
\end{center}

A more complete description of the patch augmentation algorithm follows in Section~\ref{sec:implementation} with pseudo-code and finally the experimental results are presented in Section~\ref{sec:experiments}.

\section{Implementation and Algorithm}\label{sec:implementation}
The patch augmentation algorithm has been developed in Python using Keras. Keras provides a \texttt{Sequence} class which allows you to create a generator that can be used to pass data to a network during training. Keras' \texttt{Sequence} class handles the segmentation of the data into batches, and also guarantees that each image in the training set is passed exactly once to the network during each epoch.

As can be seen in Algorithm~\ref{algo:patch-augmentation}, every image within a batch that is about to be passed to the network during training is augmented with a patch according to a user-defined probability. If this probability is set to 0, no images are altered, and the network trains with the dataset untouched. If this is set to 1 then every image within every batch is augmented. If an image is augmented, this replaces the image in the batch with the augmented sample and the same applies to its corresponding new label. Once every image in the batch is cycled through, it is passed to the network for training. By following the pseudo-code in Algorithm~\ref{algo:patch-augmentation}, it can be seen that patches are being drawn from the entire training set. This means patches could be drawn from images of the same class, or even, although highly unlikely, the same image.

\begin{algorithm}[]
\SetAlgoLined
\SetKwFunction{FRand}{Rand}
\SetKwFunction{FArea}{Area}
\SetKwFunction{FExtractPatch}{ExtractPatch}
\SetKwFunction{FPlacePatch}{PlacePatch}
\KwData{Batch of images $x$ and labels $y$}
\KwData{Training set $x'$ and labels $y'$}
\KwData{Probability $p = 0.5$}
\ForEach{Image $x_i$, label $y_i$ $\in x, y$}{
  \If{\FRand{0, 1}$<p$}
  {
    $x_r, y_r \gets$ \text{random image/label pair from } $x', y'$\;
    $x_p \gets \FExtractPatch{$x_r$}$ \;
    $x_i \gets \FPlacePatch{$x_i$, $x_p$}$ \;
    $\lambda \gets \dfrac{\FArea{$x_p$}}{\FArea{$x_i$}}$ \;
    $y_i \gets (1-\lambda)y_i + \lambda y_r$ \;
  }
}
\caption{Patch augmentation algorithm.}
\label{algo:patch-augmentation}
\end{algorithm}

Two aspects of the algorithm can be adjusted through parameters---the patch size and the probability. The patches are of a random size, defined by a minimum width/height dimension and a maximum width/height dimension. The extracted patch is then placed at a random location within the bounds of the second image. The probability parameter adjusts how many of the images are augmented---a value of 0.5 means half of the images during an epoch are augmented with patches. In our experiments, the parameters that resulted in the best overall performance were a minimum dimension of 0.3 and maximum of 0.8, and a probability of 0.9.

The augmented images' one-hot-encoded labels are generated using a linear transformation of the image pair's labels. Concretely, for a given augmented training image, the new image's label is calculated as follows (we will use the notation described in \cite{zhang2018} for consistency):

$$
\tilde{y} = (1 - \lambda)y_i + \lambda y_j
$$

The value for $\lambda$ is calculated as follows:

$$
\lambda = \frac{A{p}}{A_x}
$$

where $A_p$ denotes the area in pixels of the patch, and $A_x$ denotes the area in pixels of the image.

It follows, therefore, that:

$$
\tilde{y} = \bigg(1 - \frac{A{p}}{A_x} \bigg) y_i  +  \frac{A{p}}{A_x} y_j
$$

For example, a patch of size $200\times200$ placed within an image of size $400\times400$ would result in:

$$\lambda = \frac{A{p}}{A_x} = \frac{200 \cdot 200}{400 \cdot 400} = \frac{40 000}{160 000} = 0.25$$

Hence:

\begin{gather*}
  \tilde{y} = (1-0.25) \times [1.0,\; 0.0] + 0.25 \times [0.0,\; 1.0] \\
  \tilde{y} = [0.75,\; 0.25]
\end{gather*}

New labels are generated for each newly created, augmented image. It is a requirement for patch augmentation that the labels are one hot encoded. The approach can be generalised to any number of classes, as demonstrated later using the CIFAR-10 and CIFAR-100 data sets, which contain 10 and 100 classes respectively.

\section{Experiments and Results}\label{sec:experiments}
In our experiments to evaluate the method's effect on model accuracy, we used the CIFAR-10 and CIFAR-100 data sets. CIFAR-10 comprises 60,000 $32\times32$ pixel images across 10 classes, equally distributed. 50,000 images are used for training and 10,000 are used for testing. CIFAR-100 is identical except that is consists of 100 classes with 600 images each \cite{krizhevsky2009learning}. The ResNet networks were chosen as they are not very computationally demanding, and their saved model sizes are small and can be shared online more easily. Our hardware setup was a single workstation utilising a Titan X GPU running on Ubuntu 18.04. Due to these rather modest hardware resources, we had to consider reasonably sized networks and data sets.

All experiments, using both versions of ResNet were trained for 200 epochs, using a learning rate scheduler, adjusting from a starting learning rate of 0.001 and reducing after epoch 100, 140, 180, and 190. The loss function used throughout was categorical cross entropy in order to properly handle the one-hot-encoding categorical label vectors utilised by patch augmentation.

Our experimental results are encouraging, with patch augmentation consistently outperforming baseline accuracy on several data set and network configurations. For CIFAR-10 using ResNet20v1 we achieved a baseline accuracy of 80.86\%, which improved to 89.33\% with patch augmentation (an increase of 8.47\%). For CIFAR-10 using ResNet29v2 we achieved a baseline accuracy of 83.15\% compared to 91.19\% accuracy when using patch augmentation (an increase of 8.04\%). When compared to mixup, patch augmentation slightly outperformed it, with mixup achieving 86.62\% accuracy on CIFAR-10 using ResNet20v1. Regarding CIFAR-100, we achieved a baseline accuracy of 44.08\%, which improved to 61.41\% when using patch augmentation, an increase of 17.33\% using ResNet20v1. In turn, ResNet29v2's baseline on CIFAR-100 was measured at 52.21\%, which increased to 68.06\% when trained with patch augmentation, an increase of 15.85\% accuracy on the test set.

A summary for all results using both ResNet20v1 and ResNet29v2 trained with CIFAR-10 and CIFAR-100 can be seen in Table~\ref{tab:results-summary}.

\begin{table}[h]
  \centering
  \begin{tabular}{@{}lllll@{}}
  \toprule
  \multicolumn{2}{l}{Model/Dataset}           & No Augmentation & Fixed Patch Size   & Random Patch Size \\ \midrule
  CIFAR-10  & \multicolumn{1}{l|}{ResNet20v1} & 80.86\%         & 86.83\% (+5.97\%)  & \textbf{89.33\%} (+8.47\%)  \\
            & \multicolumn{1}{l|}{ResNet29v2} & 83.15\%         & 88.01\% (+4.86\%)  & \textbf{91.19\%} (+8.04\%)  \\ \cmidrule(l){2-5}
  CIFAR-100 & \multicolumn{1}{l|}{ResNet20v1} & 44.08\%         & 55.15\% (+11.07\%) & \textbf{61.41\%} (+17.33\%) \\
            & \multicolumn{1}{l|}{ResNet29v2} & 52.21\%         & 59.77\% (+7.56\%)  & \textbf{68.06\%} (+15.85\%) \\ \bottomrule
  \end{tabular}
  \caption{Results of the patch augmentation method, using both fixed patch sizes and random patch sizes. Random patch sizes consistently outperform fixed patch sizes. Best results in boldface.}
  \label{tab:results-summary}
\end{table}

Two main parameters can be controlled when evaluating patch augmentation's performance: namely the patch's maximum and minimum width and height (controlling the patch's area) and the probability that an image is augmented when being passed to the network for training. The parameters that were chosen are described in the sub-sections below for each data set and network architecture.

\subsection{CIFAR-10}
Baseline accuracies were measured using an identical network configuration, an identical train/test split, and an identical learning rate scheduler.
When using the ResNet20v1 network trained on CIFAR-10 a baseline performance at 80.86\% was measured. When trained with patch augmentation, using a minimum dimension of 0.3 and maximum of 0.9, and a probability of 0.9, we obtained an accuracy of 89.33\%. Switching the network architecture to ResNet29v2, a baseline accuracy of 83.15\% was recorded. With patch augmentation, accuracy increased to 91.19\%. Again this used a minimum dimension of 0.3 and maximum of 0.9, and a probability of 0.9. Slightly different parameters were chosen for CIFAR-100.

\subsection{CIFAR-100}
For CIFAR-100, we again trained both ResNet20v1 and ResNet29v2 networks. The baseline accuracy for ResNet20v1 was measured at 44.08\%. This increased to 61.41\% using patch augmentation, using a minimum dimension of 0.3 and maximum of 0.8, and a probability of 0.9. When training ResNet29v2 on CIFAR-100 a baseline accuracy was measured at 52.21\% which improved to 68.06\% when trained with patch augmentation. Baseline accuracy was measured using an identical configuration of network, as described above.

\begin{figure}[h]
  \centering
  \includegraphics[width=1.0\textwidth]{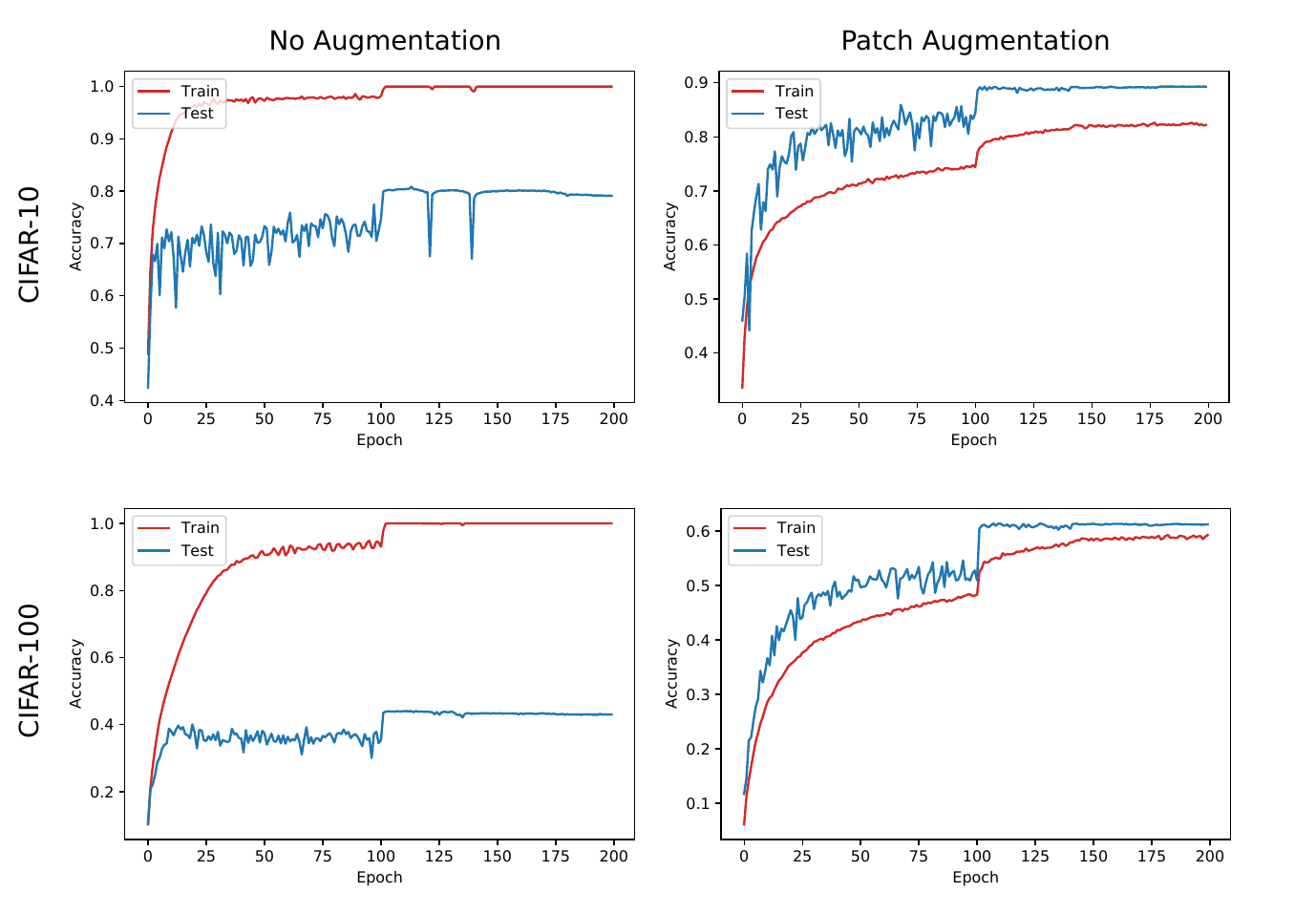}
  \caption{Accuracy comparison using ResNet20v1 trained with CIFAR-10 and CIFAR-100, with and without patch augmentation, over 200 epochs.}
  \label{fig:results-summary-plot-resnet20v1-accuracy}
\end{figure}

\begin{figure}[h]
  \centering
  \includegraphics[width=1.0\textwidth]{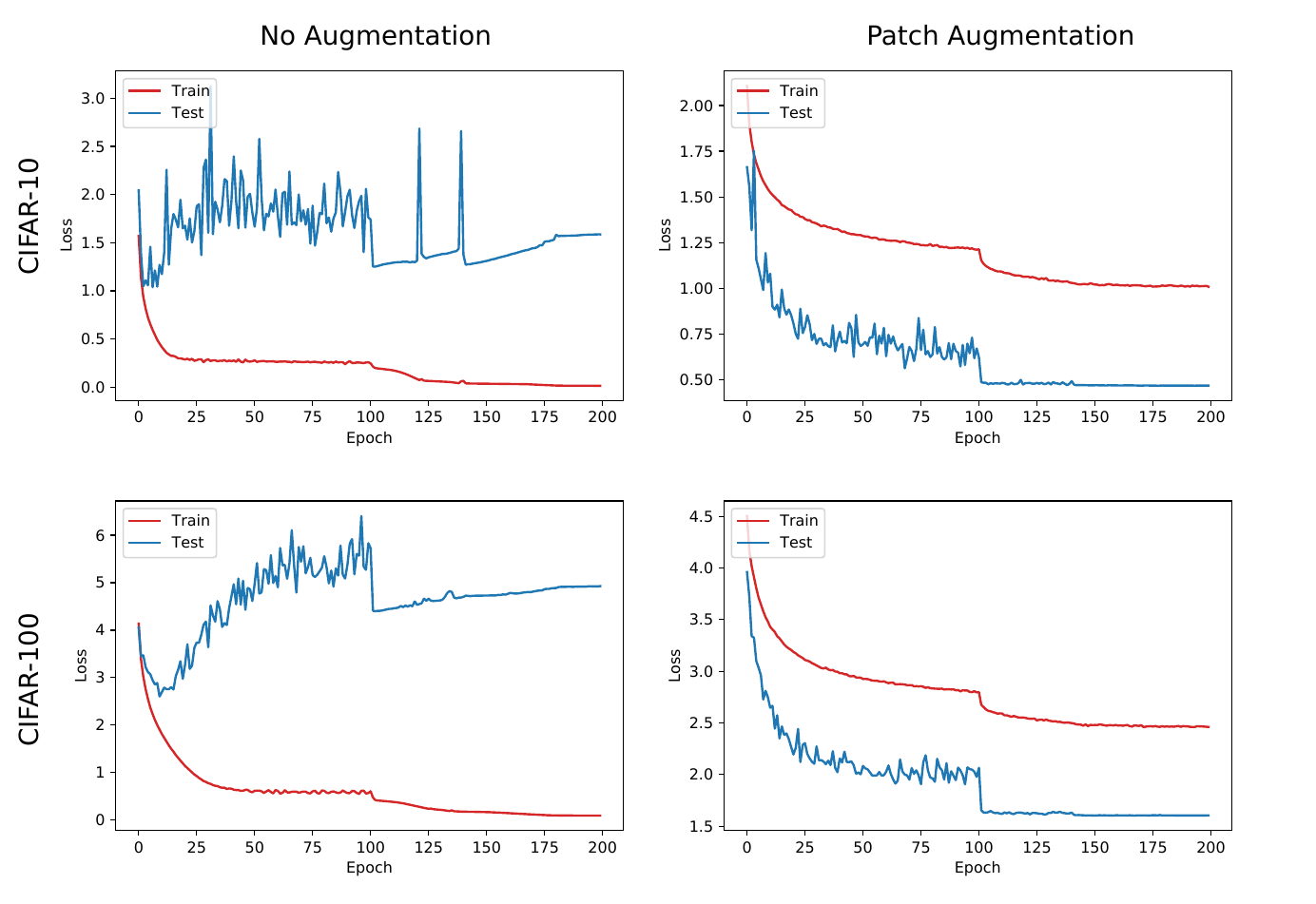}
  \caption{Loss comparison using ResNet20v1 trained with CIFAR-10 and CIFAR-100, with and without patch augmentation, over 200 epochs.}
  \label{fig:results-summary-plot-resnet20v1-loss}
\end{figure}

\begin{figure}[h]
  \centering
  \includegraphics[width=1.0\textwidth]{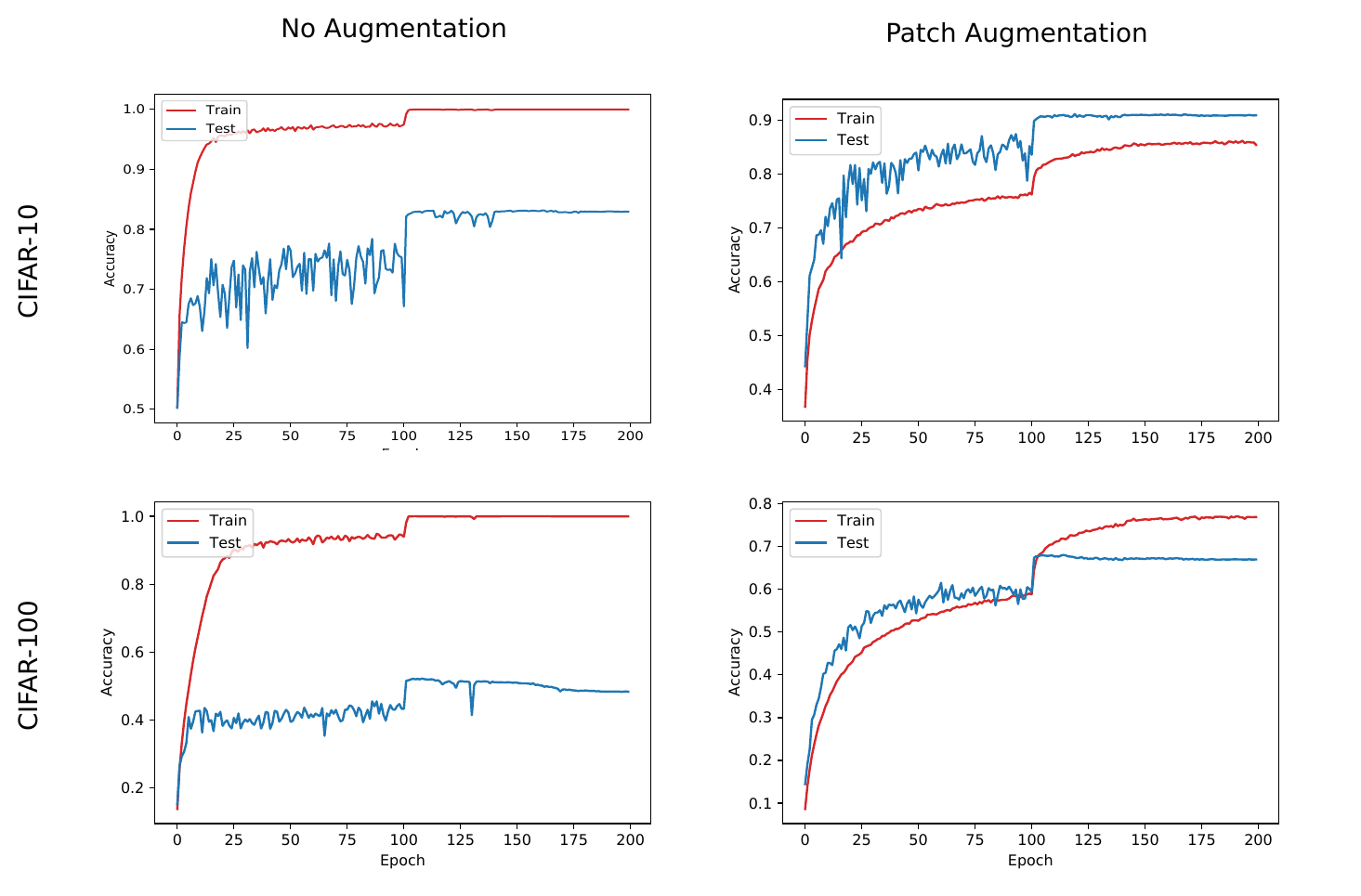}
  \caption{Accuracy comparison using ResNet29v2 trained with CIFAR-10 and CIFAR-100, with and without patch augmentation, over 200 epochs.}
\label{fig:results-summary-plot-resnet29v2-accuracy}
\end{figure}

\begin{figure}[h]
  \centering
  \includegraphics[width=1.0\textwidth]{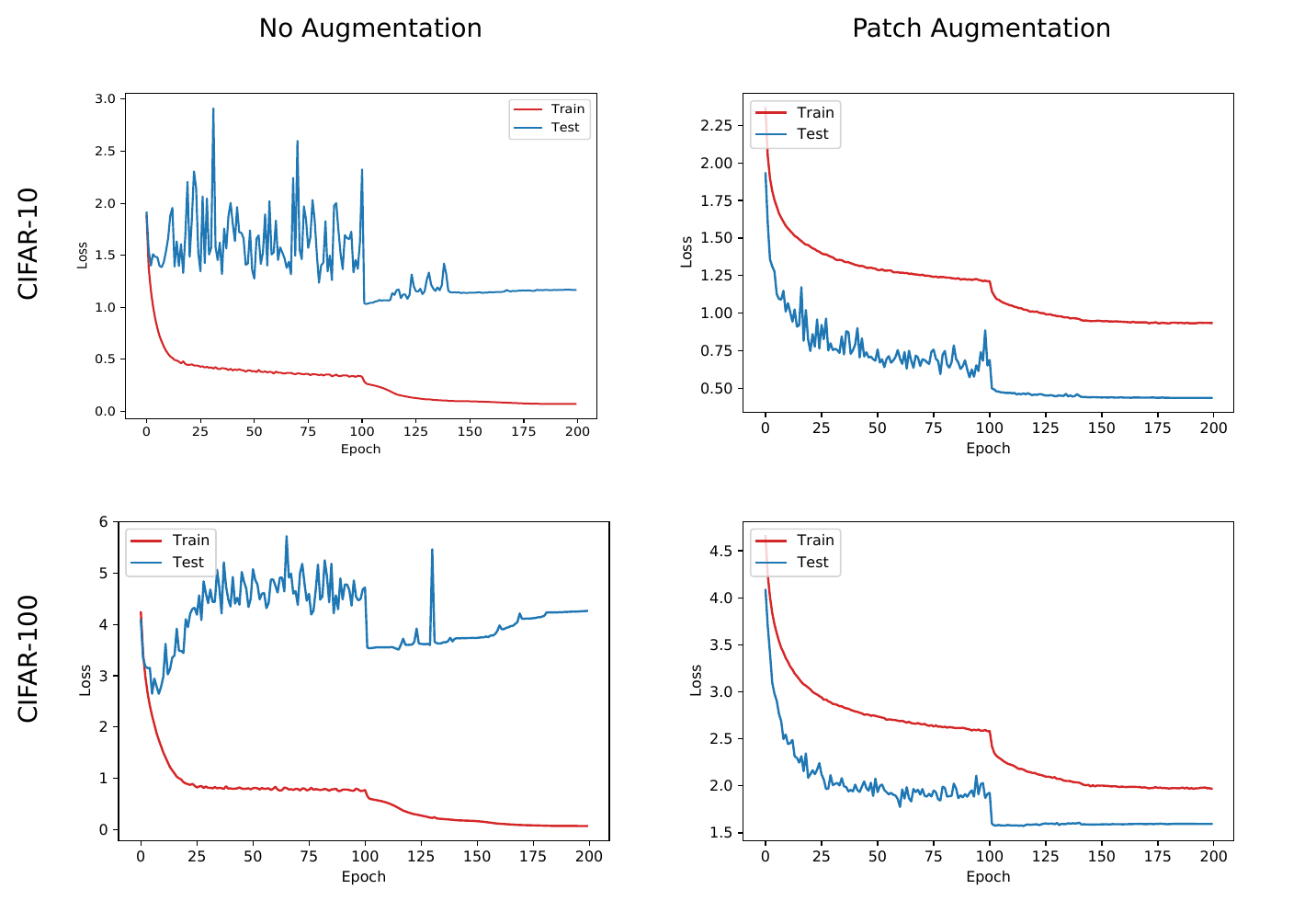}
  \caption{Loss comparison using ResNet29v2 trained with CIFAR-10 and CIFAR-100, with and without patch augmentation, over 200 epochs.}
  \label{fig:results-summary-plot-resnet29v2-loss}
\end{figure}

Side-by-side comparisons of the baseline and patch augmentation accuracies using ResNet20v1 trained with CIFAR-10 and CIFAR-100 can be seen in Figure~\ref{fig:results-summary-plot-resnet20v1-accuracy}, while a comparison of their losses can be seen in Figure~\ref{fig:results-summary-plot-resnet20v1-loss}. A comparison of the baseline and patch augmentation accuracies using ResNet29v2 trained with CIFAR-10 and CIFAR-100 can be seen in Figure~\ref{fig:results-summary-plot-resnet29v2-accuracy}, and likewise their losses can be seen in Figure~\ref{fig:results-summary-plot-resnet29v2-loss}.

\subsection{Fixed Patch Size}
In order to control from the effect of the patch size on accuracy, and to investigate whether randomly sized patches would outperform fixed sized patches, the same experiments were performed using patches of a fixed size, the results of which are presented below.

In each case the patch size was fixed at 25\% or 50\% of the area of the image depending on the data set, and the probability was set at 0.5 for all experiments. The ResNet20v1 network performance was measured at 80.86\% as a baseline, and 86.83\% using a network trained with patch augmentation, an improvement of over 5\%. ResNet29v2 network performance was 88.01\% with patch augmentation compared to 83.15\% for the baseline.

The baseline for CIFAR-100 using ResNet20v1 was 44.08\% accuracy. Using patch augmentation with fixed patch sizes, the results improved to 55.15\% , an improvement of over 10\%. The results for the patch augmentation run were obtained using a probability of 0.5 and a patch area of 50\%, as opposed to 25\% for the CIFAR-10 runs. When using ResNet29v2, we saw the baseline for CIFAR-100 of 52.21\%. However, CIFAR-100 with patch augmentation improved to 59.77\%, an increase in accuracy of over 7\% from the baseline. See Table~\ref{tab:results-summary} for these results compared to those using randomly sized patches.

While the accuracies obtained using a fixed patch size outperformed the baseline values in all experiments, the best results were consistently obtained using randomly sized patches. This is likely due to the increased variance of the datasets produced by the randomised approach.

\section{Robustness to Adversarial Attacks}
In order to test the robustness of the networks trained with patch augmentation to adversarial attacks, we used the \textit{CleverHans} library \cite{papernot2018cleverhans}.

CleverHans implements a number of adversarial attack creation algorithms. These attacks accept a model and a test set as input, and they return the corresponding adversarial examples. A number of attacks are available, however we tested the trained network using the Fast Gradient Sign Method (FGSM) \cite{goodfellow2014explaining}. In this white-box attack, where access to the trained model is required to generate the attacks, the network's gradients are used to create the adversarial example that will trick the network. For a given input image, $x$, FGSM uses the gradients of the loss w.r.t. the input image to create a new image, $\tilde{x}$, that maximises the loss \cite{tensorflowdocs}:

$$
\tilde{x} = x + \epsilon \cdot \text{sign}( \nabla_x J(\theta, x, y) )
$$

where $\epsilon$ controls the magnitude of the perturbation---in our case, this was set to 0.001 and 0.03 to see the effect of the perturbation change. Larger values of $\epsilon$ means the network is more likely to misclassify $\tilde{x}$, however this also means that it is more detectable by a human. We benchmarked the robustness of our method to adversarial attack using v3.0.1 of CleverHans. On a test set modified by the Fast Gradient Sign Method with a max-norm $\epsilon$ of 0.001, we obtained a test set accuracy of (on the first 1,000 images in the CIFAR-10 test set) of 64.3\% accuracy using the non-augmented model, versus 72.5\% accuracy using the model trained with patch augmentation (See Table~\ref{tab:adversarial-summary}).

\begin{table}[h]
  \centering
  \begin{tabular}{@{}lll@{}}
  \toprule
  Adversarial Attack & No Augmentation  & Patch Augmentation  \\ \midrule
  FGSM ($\epsilon$ = 0.001)   & 64.3\%  & \textbf{72.5\%}     \\
  FGSM ($\epsilon$ = 0.03)    & 13.8\%  & \textbf{20.1\%}     \\ \bottomrule
  \end{tabular}
  \caption{Summary of trained model accuracies for FGSM generated adversarial examples (ResNet20v1/CIFAR-10). Model trained using fixed patch size of 25\% and a probability of 0.5.}
  \label{tab:adversarial-summary}
\end{table}

When increasing $\epsilon$ to 0.03, accuracy degraded significantly, however the network trained with patch augmentation still outperformed the non-augmented approach with an accuracy of 20.1\% compared to 13.8\% for the standard model.

\section{Conclusions}
Data augmentation is an integral part of deep learning, and is used almost by default for any neural network based model learning on image data. It is used to train more general models, avoid over-fitting, and avoid memorisation.

In this paper we have provided a data-independent approach that increases classification accuracy in a number of scenarios, but has also been shown to strengthen networks against adversarial attacks. In terms of accuracy, we have seen that patch augmentation increased the accuracy of neural networks on some common data sets when compared to baseline models without any augmentation applied. Accuracy increased by over 8\% for CIFAR-10 and over 17\% for CIFAR-100. We have also seen that networks trained using patch augmentation are more robust to adversarial examples and attacks generated with the Fast Gradient Sign Method. This is of particular importance in artificial intelligence research, due to recent papers highlighting how fragile some normally very well performing networks seem to be.

Further work may be needed to ascertain the best parameter choices for the algorithm. We plan to explore the effect of augmentation on the patch data itself, such as slight rotations or the addition of transparency.  Also, we will experiment with the placement of patches wholly within the host image, or allowing patches to overlap beyond the border of the host image (in effect being cropped out of the newly augmented image). Future work will concentrate on finding optimal approaches, and there is much scope here for further development.

Last, the technique will be fully integrated in to the Augmentor software library as a standard feature of the package, so that patch augmentation can be applied to existing machine learning training pipelines conveniently.

%
\section*{Conflict of interest}
The authors declare that they have no conflict of interest.


\bibliography{references}   

\begin{thebibliography}{10}
\providecommand{\url}[1]{{#1}}
\providecommand{\urlprefix}{URL }
\expandafter\ifx\csname urlstyle\endcsname\relax
  \providecommand{\doi}[1]{DOI~\discretionary{}{}{}#1}\else
  \providecommand{\doi}{DOI~\discretionary{}{}{}\begingroup
  \urlstyle{rm}\Url}\fi

\bibitem{bloice2019augmentorbio}
Bloice, M.D., Roth, P.M., Holzinger, A.: {Biomedical image augmentation using
  Augmentor}.
\newblock Bioinformatics \textbf{35}(1), 4522--4524 (2019)

\bibitem{cubuk2018auto}
Cubuk, E.D., Zoph, B., Mane, D., Vasudevan, V., Le, Q.V.: Autoaugment: Learning
  augmentation policies from data.
\newblock arXiv:1805.09501  (2018)

\bibitem{eykholt2018robust}
Eykholt, K., Evtimov, I., Fernandes, E., Li, B., Rahmati, A., Xiao, C.,
  Prakash, A., Kohno, T., Song, D.: Robust physical-world attacks on deep
  learning visual classification.
\newblock In: Proceedings of the IEEE Conference on Computer Vision and Pattern
  Recognition (CVPR 2018), pp. 1625--1634 (2018)

\bibitem{finlayson2019}
Finlayson, S.G., Bowers, J.D., Ito, J., Zittrain, J.L., Beam, A.L., Kohane,
  I.S.: Adversarial attacks on medical machine learning.
\newblock Science \textbf{363}(6433), 1287--1289 (2019)

\bibitem{ghorbani2019interpretation}
Ghorbani, A., Abid, A., Zou, J.: Interpretation of neural networks is fragile.
\newblock In: Proceedings of the AAAI Conference on Artificial Intelligence,
  vol.~33, pp. 3681--3688 (2019)

\bibitem{goodfellow2014explaining}
Goodfellow, I.J., Shlens, J., Szegedy, C.: Explaining and harnessing
  adversarial examples.
\newblock arXiv preprint arXiv:1412.6572  (2014)

\bibitem{heaven2019deep}
Heaven, D.: {Why deep-learning AIs are so easy to fool}.
\newblock Nature \textbf{574}, 163--166 (2019)

\bibitem{ho2019population}
Ho, D., Liang, E., Stoica, I., Abbeel, P., Chen, X.: Population based
  augmentation: Efficient learning of augmentation policy schedules.
\newblock arXiv:1905.05393  (2019)

\bibitem{krizhevsky2009learning}
Krizhevsky, A.: Learning multiple layers of features from tiny images.
\newblock Tech. rep. (2009)

\bibitem{lecun1998gradient}
LeCun, Y., Bottou, L., Bengio, Y., Haffner, P.: Gradient-based learning applied
  to document recognition.
\newblock Proceedings of the IEEE \textbf{86}(11), 2278--2324 (1998)

\bibitem{papernot2018cleverhans}
Papernot, N., Faghri, F., Carlini, N., Goodfellow, I., Feinman, R., Kurakin,
  A., Xie, C., Sharma, Y., Brown, T., Roy, A., Matyasko, A., Behzadan, V.,
  Hambardzumyan, K., Zhang, Z., Juang, Y.L., Li, Z., Sheatsley, R., Garg, A.,
  Uesato, J., Gierke, W., Dong, Y., Berthelot, D., Hendricks, P., Rauber, J.,
  Long, R.: Technical report on the cleverhans v2.1.0 adversarial examples
  library.
\newblock arXiv preprint arXiv:1610.00768  (2018)

\bibitem{papernot2016transferability}
Papernot, N., McDaniel, P., Goodfellow, I.: Transferability in machine
  learning: from phenomena to black-box attacks using adversarial samples.
\newblock arXiv:1605.07277  (2016)

\bibitem{su2019pixel}
Su, J., Vargas, D.V., Sakurai, K.: One pixel attack for fooling deep neural
  networks.
\newblock IEEE Transactions on Evolutionary Computation  (2019)

\bibitem{szegedy2013adversarial}
Szegedy, C., Zaremba, W., Sutskever, I., Bruna, J., Erhan, D., Goodfellow, I.,
  Fergus, R.: Intriguing properties of neural networks.
\newblock arXiv:1312.6199  (2013)

\bibitem{tensorflowdocs}
{TensorFlow Documentation}: {Adversarial example using FGSM}.
\newblock
  \url{https://www.tensorflow.org/tutorials/generative/adversarial_fgsm}.
\newblock Accessed: 2019-11-05

\bibitem{zhang2018}
Zhang, H., Cisse, M., Dauphin, Y.N., Lopez-Paz, D.: mixup: Beyond empirical
  risk minimization.
\newblock arXiv:1710.09412  (2017)

\end{thebibliography}
\bibliographystyle{spmpsci}      

\end{document}